\pdfoutput=1
\documentclass[11pt]{article}

\usepackage[final]{acl}
\usepackage{graphicx}

\usepackage{times}
\usepackage{latexsym}

\usepackage[T1]{fontenc}
\usepackage[utf8]{inputenc}

\usepackage{microtype}

\usepackage{algorithm}
\usepackage{amsmath}
\usepackage{algorithmic}

\title{An Interactive Framework for Profiling News Media Sources}

\author{Nikhil Mehta \\ Department of Computer Science \\ Purdue University \\ West Lafayette, IN 47907 \\ {\tt mehta52@purdue.edu} \\\And Dan Goldwasser \\ Department of Computer Science \\ Purdue University \\ West Lafayette, IN 47907 \\ {\tt dgoldwas@purdue.edu} \\}

\begin{document}
\maketitle
\begin{abstract}
The recent rise of social media has led to the spread of large amounts of fake and biased news, content published with the intent to sway beliefs. While detecting and profiling the sources that spread this news is important to maintain a healthy society, it is challenging for automated systems.

In this paper, we propose an interactive framework for news media profiling. It combines the strengths of graph based news media profiling models, Pre-trained Large Language Models, and human insight, to characterize the social context on social media. Experimental results show that with as little as 5 human interactions, our framework can rapidly detect fake and biased news media, even in the most challenging settings of emerging news events, where test data is unseen. 
\end{abstract}

\section{Introduction}
\label{sec:intro}
The recent rise of social media has enabled information to spread at a rapid pace, having the potential to very quickly impact a large number of people, especially during key events, such as political elections \cite{vosoughi2018spread}. While this rise of social media has many benefits, one downside is that harmful information, i.e. fake or politically biased news, can also spread rapidly, affecting people's perspectives. Thus, detecting it is important.

While one approach is to fact-check or detect the bias of all content on social media (i.e. Twitter), another is to focus on the source, and ask: \textit{Is this source factual or politically biased?} This task, \textbf{profiling news media sources}, which we focus on, can scale better, as often times a large amount of the content sources' publish have the same factuality/political bias as the source itself. We model this on a 3-point scale: \textit{high, low, and mixed} factuality, and \textit{left, center, and right} bias. Details: App.~\ref{sec:source_task_importance}.

% , and new ones arising daily
Even at the source level, it is difficult for humans to profile all news content, due to the large number of sources online. Further, this task is still challenging for AI systems \cite{baly:2018:EMNLP2018, baly:2020:ACL2020, mehta2022tackling}, especially in the \textit{emerging events settings}, when the system is tested on its \textbf{ability to adapt to new events}, consisting of new sources, content they generate, and social media users engaging with them that were not seen at training time \cite{yuan2020early}. For example, in a graph framework, test set nodes are not connected to training set nodes. In these settings, Large Language Models (LLMs) also struggle, even when enhanced with extra knowledge \cite{whitehouse2022evaluation}.

Due to the struggles of AI systems to automatically profile news media, in this paper we propose a different, interactive approach, for this task. We are inspired by recent results~\cite{
cinelli2021echo, doi:10.1073/pnas.1517441113} showing that misinformation and highly biased content tends to spread in closely knit communities on social media. This leads us to ask, whether better modeling of the social media relationships that underlie content spread would improve our ability to profile the content itself. Specifically, we hypothesize that users on social media form \textit{information communities}, or groups of users, where certain themes circulate more in some communities vs. others. If we can identify these themes and use them to \textbf{identify information communities}, then we can better profile the content discussed by the communities. For example, a user joining a left-leaning community, is more likely to be left-leaning, and so is the content they share. Then, if that user shares content from a source, that source is also more likely to be left-leaning.

\begin{figure*}[t!]
  \centering
  \includegraphics[scale=0.33]{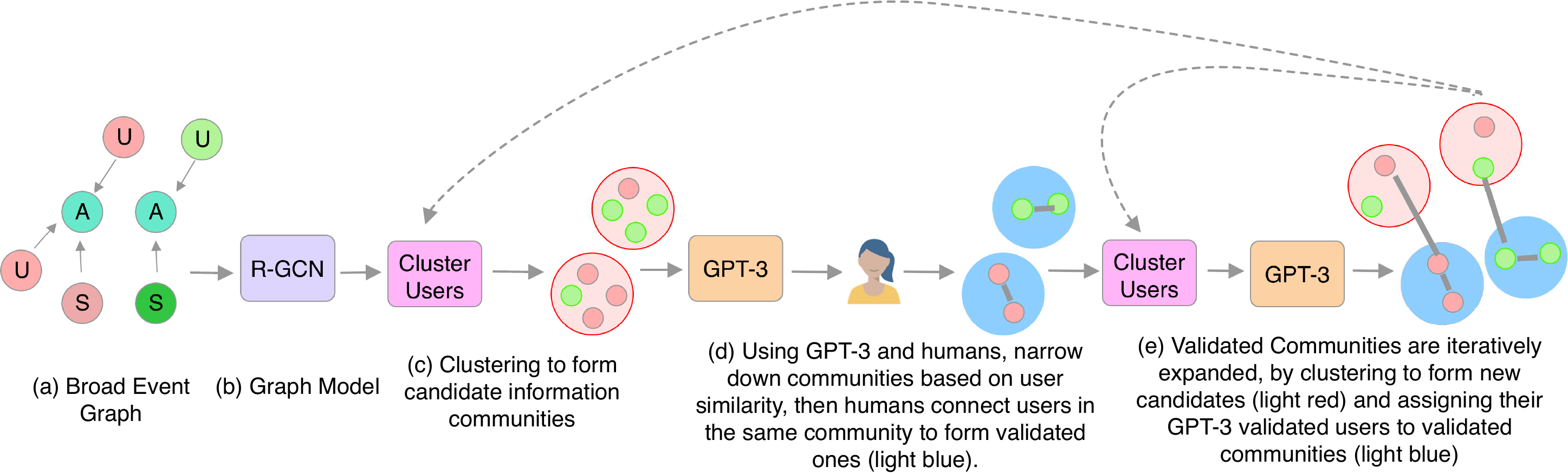}
\caption{\small Our framework overview: \textbf{Using Trained Graph Models, Large Language Models (GPT-3) and Human Interaction to Form Information Communities for News Media Profiling}. (Key: U = Users, A = Articles, S = Sources, Light Red Background = Candidate information Communities, Light Blue Background = Validated Information Communities). From the learned graph model (b), we find candidate information communities through k-means clustering. Using a LLM, GPT-3, we form a textual representation of the information community by summarizing its users, and then ask humans to narrow down the community based on the user summaries (d), forming smaller, validated communities, whose users are then connected to each other. We then expand the validated communities, by again model clustering users, forming user summaries, but this time asking GPT-3 to place or not palce the users into validated communities, which can be done repeatedly (e). This entire process (c-e) can repeat, starting with clustering of unassigned users (c), to form more validated communities, which can be expanded further.}
\label{fig:overview}
\vspace{-10pt}
\end{figure*}

In the settings of emerging events, we must be able to characterize and form the information communities quickly, without using labeled training data, so we can rapidly detect fake/biased news sources on unseen data.  Unfortunately, using only information on social media to form information communities involves complex reasoning on unseen data, and is thus challenging for automated systems, even LLMs (as we experimentally show). Thus, instead, in this paper, we propose an \textbf{interactive learning framework}, to form information communities. We take advantage of minimal human interactions, combining it with the strengths of news media profiling graph models and LLMs, to form the communities. We do this in an iterative process, showing how less than 5 human interactions, which do not label any additional data, can lead to significant improvements in both fake news and bias source detection, even in the challenging weakly supervised emerging events settings.

Specifically, to form the communities, we first form initial candidate communities by clustering social media users based on their graph model embeddings, capturing model beliefs. As news media profiling is difficult, these candidate information communities are likely to be imperfect, i.e., some users are likely to be inconsistent with their community assignment. Thus, these assignments  need to be validated, by examining their actual information preferences and the content they generate, rather than using only relational information. This would help ensure that only users that are actually similar to each other (i.e. have similar perspectives on similar content) make up each community. 
We use LLMs and human interaction to do this validation. First, we summarize each user in the community using LLM's, mapping their profile to a short statment capturing their views. This representation facilitates the human validation step, in which we ask humans to determine which users are similar, and by that form validated communities (experiments show this is hard for LLMs, but simple for humans). We then expand the validated communities, adding new users to them, based on model clustering. These assignments are again validated based on users' summaries, but this time using LLMs to compare new users to users in human validated communities (we use human interactions as training examples for few-shot user similarity detection, a simpler LLM task). For users not assigned to validated communities, the above process is repeated, expanding the \# of communities. Fig.~\ref{fig:overview} shows an overview. In short, humans interact to help form initial communities, which are then automatically expanded using graph and LLM similarity.

In summary, we make the following contributions: \textbf{(1)}: We formulate the task of interactive news media profiling, by presenting a framework to build information communities. \textbf{(2)} We take advantage of graph model, LLM, and human knowledge to perform this interactive task rapidly. \textbf{(3)} We evaluate on one of the most challenging news media profiling settings, emerging news events, showing how our interactive framework leads to performance improvements with less than 5 human interactions. More generally, our framework can be used to rapidly profile news media, without any additional labeling and minimal human effort. 

Sec.~\ref{sec:model} describes our graph, Sec.~\ref{sec:interactions} our interactive framework, Sec.~\ref{sec:experiments} results, and Sec.~\ref{sec:discussion} analysis.

 \section{Related Work}
 \label{sec:related_work}
 Over the last few years, there has been a large interest in profiling news media. \citeauthor{baly:2018:EMNLP2018} and \citeauthor{sakketou2022factoid} proposed datasets for political bias detection, while \cite{li2019encoding,liu2021mitigating, liu2022quantifying} study methods for it. Further, fake news detection has also been a hot research topic, studied in graphs \cite{nguyen2020fang, mehta2022tackling, yang2023entity}, cross-domain \cite{huang2021dafd, ENDEF, zhu2022memory, mosallanezhad2022domain}, and low-resource \cite{lin2022detect} settings, amongst others. 

We focus specifically on news media profiling in the emerging events setting, which is extremely challenging, as test data is unseen and does not interact with any train data. Thus, this setting is ideal for us to measure the benefits of human interactions. It also has received a lot of recent attention due to its' challenge \cite{liu2018early, li2022adadebunk}, and we hypothesize that some of these methods can be combined with our work.

Large Language Models (LLMs) have been applied to many tasks, as they can capture lots of knowledge \cite{qinChatGPT}. However, LLMs still cannot reason well, and thus struggle on harder tasks like fake news detection \cite{whitehouse2022evaluation}. We instead use LLM's successful properties to amplify the impact of human interactions.

Interactive ML has also been studied and applied to many tasks \cite{wu2022survey, dalvi2022towards, kwonreward, Ramamurthy2022IsRL,pacheco2022holistic,pacheco2023interactive, mehta2023interactively}. Building information communities is also a popular research area, whether it be through embeddings or DNN modeling approaches \cite{cavallari2017learning, su2022comprehensive}. Prior work also shows misinformation spreads in communities \cite{bessi2016homophily}. We propose to interactively build information communities, by humans interacting with LLMs and graph models. We discuss more related works in App.~\ref{appendix:related_cont}.

 \section{Graph Overview}
 \label{sec:model}
 In this paper, we focus on detecting the political bias and factuality of news media sources, which we call \textbf{news media profiling}. We model sources for factuality and political bias on a 3-point scale: \textit{low}, \textit{mixed}, and \textit{high} factuality, or \textit{left}, \textit{center}, and \textit{right} political bias. More details about this task setup and its importance are in \cite{baly:2020:ACL2020} and Appendix~\ref{sec:source_task_importance}.

We use the public graph-based social media analysis model from \citeauthor{mehta2022tackling}\footnote{\url{https://github.com/hockeybro12/FakeNews_Inference_Operators}}, which they trained for fake news source detection. As we also evaluate news source bias detection, we train the graph for both objectives, on Twitter data. We refer to \citeauthor{mehta2022tackling} for the details of this graph model, but briefly explain it here. Sec.~\ref{sec:interactions} explains our interactive protocol for identifying information communities.

The model uses a heterogeneous graph, encoded using Relational Graph Conv. Networks (R-GCN), to capture the relationships between sources, articles, and users. Based on the R-GCN representations, factuality and bias of news sources can be predicted. R-GCNs allow us to better capture relationships in the graph, such as a source represented in part by the users that follow it (which are also represented by their relationships to other nodes).

\textbf{Graph Creation using Twitter Social Context:}
Our graph (Fig.\ref{fig:overview}a, \citeauthor{mehta2022tackling}) consists of 3 node types: News Sources ($S$), Articles ($A$) they publish, and Twitter Users ($U$) that interact with sources and articles. We connect sources to articles they publish, via edges. Further, users are connected to sources and other users they follow, and articles they propagate (retweet or tweet the link of). The Twitter users provide the social information in the graph, which we later aim to better learn.

\textbf{Graph Training using Social Context:}
Similar to \citeauthor{mehta2022tackling}, we train a R-GCN \cite{schlichtkrull2018modeling} to learn the graph. We train the classification objective of both fake news source detection and news source bias detection, using a separate Fully Connected (FC) layer for each, optimizing them jointly by summing the losses. Once the model is trained, we can use it to obtain meaningful node embeddings for every node in the graph, and profile news sources. More details: App.~\ref{appendix:model_training}.
 
 \section{Interactive Approach}
 \label{sec:interactions}
 While the above graph-based model proposed by \citeauthor{mehta2022tackling} achieves strong performance on fake news source detection when evaluated in transductive settings (test data seen at training time), it struggles in the fully inductive settings (test data unseen), and in general performs well below human baselines. We thus propose an interactive approach, combining the strengths of graph models, large language models (LLMs i.e. GPT-3), and humans, to profile content on social media better. 

Our approach hinges on the fact that \textbf{if we can better model user preferences and thus user similarity, we can better model the content they propagate}. This is because, similar users are likely to have similar interests, and thus share similar content, which in turn is likely to have similar levels of factuality/bias. For example, a group of users sharing content in support of lowering taxes and decreasing regulations are more likely to be right-biased (i.e. Republican) vs. left-biased (i.e. Democrat), and thus any source content they share is also more likely to be right-biased. Thus, we hypothesize that the larger the groups of users with similar content preferences we can form, the higher our performance is likely to be. Further, if we \textbf{explicitly create new graph edges between these similar groups of users}, this information will flow to other users not part of these groups, and eventually news sources, increasing classification performance.

Fortunately, while modeling user content preferences solely through AI models like LLMs is difficult \cite{whitehouse2022evaluation}, humans can quickly determine if two users are similar, forming an initial group. Then, we hypothesize, that LLMs can be prompted using the human insight to extend the group, by asking them if other users are similar in the same ways. Thus, in this paper, we propose an \textbf{interactive framework}, taking advantage of human and LLM strengths to better model user content preferences, and improve media profiling.

Specifically, in this Sec., we discuss the interactive approach we propose to form these groups of similar users, or \textit{information communities}. We take advantage of the strengths of trained graph model knowledge, LLM knowledge (GPT-3), and human insight to design an iterative, interactive approach: We first use the graph models' learned user similarities to form initial candidate information communities (Sec.~\ref{sec:graph_comms}), which are summarized by LLM's (Sec.~\ref{sec:characterize_gpt3}), and then validated by a human interactor (Sec.~\ref{sec:inital_human_interaction}). These validated communities are then expanded upon to include more users in Sec.~\ref{sec:expand_gpt3}, by again using graph model knowledge and LLM's. However, this time LLM's are prompted based on the human interaction, to validate user assignments to communities. This expansion step can be done iteratively, i.e. assigning additional users to validated communities. Once enough users are not assigned to existing communities, we form an additional set of human validated communities, repeating the the above process from Sec.~\ref{sec:graph_comms}-Sec.~\ref{sec:expand_gpt3}, and the number of total validated communities is increased (Sec.~\ref{sec:repeat_comm_expand}). Finally, once enough information communities are formed, we can then learn them (Sec~\ref{sec:learning_graph_embeddings}), updating graph model parameters.

We later show that through this iterative process, minimal human interactions can lead to significant performance improvements for news media profiling. Fig.~\ref{fig:overview} and Alg.~\ref{alg:soliciting_interactions} shows an overview.

\begin{algorithm}
\caption{\textit{Our Interactive Framework to Find Validated Information Communities}}
\begin{algorithmic}[1]
  \small
    \STATE \textbf{Input:} $U$ (Users), $U_E$ (Graph User Embeddings), V (empty list to store validated information communities)
  \STATE \textbf{Output:} $V$ (Validated Communities)

\STATE {\scriptsize{\fontfamily{pcr}\selectfont Iteratively find information communities}}
\WHILE {not converged}
\STATE $c_{1...k} = \text{k-means}(U_E)$ {\scriptsize{\fontfamily{pcr}\selectfont K-means Cluster all Users based on Graph User Embeddings}}
\STATE $c_1, c_2 = max_2(\text{purity}(c_{1...k}))$ {\scriptsize{\fontfamily{pcr}\selectfont Choose the highest purity clusters, that discuss the same entity}}
\STATE $s_1, s_2 = \text{GPT-3 summarize}(c_1, c_2)$ {\scriptsize{\fontfamily{pcr}\selectfont Use GPT-3 to summarize the users in each cluster}}
\STATE $V.\text{append}([\text{human}(s_1), \text{human}(s_2)])$ {\scriptsize{\fontfamily{pcr}\selectfont Validate clusters using humans to form validated communities}}

\STATE {\scriptsize{\fontfamily{pcr}\selectfont Now, iteratively expand each validated community}}
\WHILE {not converged}
   \STATE $c_{1...k} = \text{k-means}(U_E)$ {\scriptsize{\fontfamily{pcr}\selectfont Again Cluster all Users based on Graph User Embeddings}}

   \STATE $c_{1...k} = \text{KNN}(c_{1...k}, V)$ {\scriptsize{\fontfamily{pcr}\selectfont For each cluster, find the $m$ nearest neighbors to each validated community, that's our new cluster}}
   
   \STATE $V = (\text{GPT-3}(c_{1...k}, V))$ {\scriptsize{\fontfamily{pcr}\selectfont For each cluster, ask GPT-3 to assign or not assign users to validated clusters, expanding them}}
    
\ENDWHILE
\ENDWHILE

\RETURN $V$ (Validated Information Communities)

 \end{algorithmic}
\label{alg:soliciting_interactions}
\vspace{-5pt}
\end{algorithm}
\vspace{-5pt}

\subsection{Initial Communities from Graph Model}
\label{sec:graph_comms}
The first step in our process of forming information communities of similar users is forming candidate ones. For this, we use learned graph model knowledge, and $k$-means cluster all graph user node embeddings, as similar nodes will be part of the same cluster (and thus community). We keep the two (determined empirically using the dev. set) highest purity clusters, as the model is likely most confident about them, since it predicts similar users as having the same labels. To compute purity, each cluster is assigned to a class based on the most predicted user label in that cluster, and then the accuracy of this is measured. To get predicted user labels, we assign each user the label of the most common source they follow + article they tweet. 

Since these communities are formed using graph learned relationships, they are likely imperfect, and should be analyzed to form better communities. Thus, we ask humans to analyze them. However, as the communities have a lot of users which would require a lot of interactions, we narrow them down. We only keep users that discuss the most common entity mentioned in the community, as discussion around this entity can represent the community's perspective. To do this, we run an Entity Recognition system \cite{akbik2019flair} on the articles each community user tweets, keeping users if they tweet an article containing the most frequent entity in the community. We now have initial model predicted information communities of users that discuss the same entity, and thus likely the same event.

\subsection{Characterizing Users Using GPT-3}
\label{sec:characterize_gpt3}
Before asking humans to validate communities based on user similarity, we form a textual representation for each user, that can be analyzed. While not essential, this representation captures relevant content and user preferences, making human interaction easier. To form it, we use LLMs (GPT-3), prompting them to create user summaries, as they have historically done well on this task \cite{qinChatGPT}. The summary for each user is formed based on their Twitter profile and a sample of their tweets related to the entity. An ex. of the prompt we designed is shown below in Tab.~\ref{tab:summary_prompt} and Fig.~\ref{fig:user_summary}.

\begin{table}[h!]
\begin{center}
\begin{tabular}{|c|c|}
  \hline
  {\textbf{\small Format}} & {\textbf{\small Language}}\\
 \hline
  \small Question & \small \begin{tabular}[x]{@{}c@{}}What is the user discussing \\and what is their perspective?\end{tabular} \\ 
  \hline 
  \small Text & \small \begin{tabular}[x]{@{}c@{}}Bio: ... Tweet 1: ... Tweet 2:... \\ Summary: \end{tabular} \\
  \hline 
  \small Output & \small The user is discussing... \\
 \hline
\end{tabular}
\caption{\small The question, text, and output format expected from GPT-3 in the prompt to create user summaries. }
\label{tab:summary_prompt}
\end{center}

\vspace{-20pt}
\end{table}

\subsection{Human Interaction to Form Community}
\label{sec:inital_human_interaction}
Based on the GPT-3 summaries of each user in the communities, we ask a human interactor to tighten each community, and only keep similar users. For this, \textbf{humans read the summaries, analyzing user perspective towards the entity}. We say users have the same perspective if they discuss the same entity in a similar way (i.e. all are against BLM protests). 

To make this analysis easier for humans, we also provide humans with an LLM's opinion on which users are similar. While it is likely incorrect, as LLMs can't reason well about user similarity on unseen topics (see Sec.~\ref{sec:importance_of_humans}), it can help humans make their decision quicker. To get it, we feed all user summaries to a dialogue LLM, Chat-GPT\footnote{\url{https://chat.openai.com/chat}}, asking it: \textit{Which users have the same perspective?} We use Chat-GPT instead of GPT-3 as it is better suited to respond without being prompted, and it is hard to create a general enough prompt for this.

Chat-GPT responds with a list of users that it thinks have the same perspectives, for ex.: \textit{User a, b, c, d discuss ... while e, f discuss ...}, which the human then reads (along with the summaries) and uses to form a human validated community. For ex., the human can decide users $(a, b, c)$ are in the same community, where $d$ was thought to be part of it by Chat-GPT, but not by the human. An ex. of the exact text humans read is in Fig.~\ref{fig:chat_gpt_output}.

\subsection{Automatically Expanding Communities}
\label{sec:expand_gpt3}
So far, we have formed small, human validated information communities, each of a single perspective. Now, we amplify this human interaction by expanding these communities, while maintaining the same perspective. We do this by identifying other users that have the same perspective and adding them to the community. These larger communities can then be used to to profile news media better.

We first connect the users in each validated community to each other, which changes their and other users' graph embeddings, without any training. We then $k$-means cluster all user embeddings, ignoring any users already considered (to avoid redundancy). This forms $k$ new, unique clusters, based on learned graph model knowledge. We hypothesize that if we can accurately map some users from each of these different clusters to validated communities, we would have a lot more information about each of those clusters, which could help news media profiling. (Aside: an alternative way to expand communities, which we do not pursue, is by assigning users that have similar embeddings as validated communities to them. However, this would just reinforce existing model predictions, as the model already believes these users similar, and thus likely not lead to better news media profiling.)

To map users from clusters to validated communities, for each cluster, we keep the top $m$ users that have similar embeddings to the centroids of each validated community, as these are the most likely users to belong to the community. These $m$ users are now the candidate users for expansion into the validated community.

As this user to community assignment is based only on graph knowledge, it may be imperfect, so we hypothesize to use LLMs to clean it up. While LLMs cannot reason about community assignments on unseen news events (we experimentally show this), which is why we used humans in Sec.~\ref{sec:inital_human_interaction}, we hypothesize that if prompted appropriately, they can compare user summaries on a topic, which the human communities are already centered around. Building on this, we few-shot prompt LLMs, asking them to identify user similarity and determine which of the $m$ new users should be part of the validated community. To do this well, 
we prompt the LLM using a training example, created automatically from the human validated community. In it, community assignments humans chose when interacting are positive examples, while ones humans rejected are negative. LLMs now just have to make similar assignments as humans (i.e. determine if the new users are more similar to human accepted or rejected ones), a much simpler task. Tab.~\ref{tab:expansion_prompt} shows an ex. of the test prompt for the community with users $(a, b)$, where user $a$ is assigned to the community and $b$ is not. The same prompt format is used for the training example, except the summaries and assignments are provided based on the human interaction. Fig.~\ref{fig:user_comm_membership} shows the full prompt.

\begin{table}[h!]
\begin{center}
\begin{tabular}{|c|c|}
  \hline
  {\textbf{\small Format}} & {\textbf{\small Language}}\\
 \hline
  \small Question & \small \begin{tabular}[x]{@{}c@{}}Which users have the same \\ perspective?\end{tabular} \\ 
  \hline 
  \small Text & \small \begin{tabular}[x]{@{}c@{}} User A Summary: ... \\ User B Summary: ... \\ Related Users;;;;Not Related Users: \end{tabular} \\
  \hline 
  \small Output & \small \begin{tabular}[x]{@{}c@{}}User A;;;;User B\end{tabular}  \\
 \hline
\end{tabular}
\caption{\small The question, text, and output format expected from GPT-3 in the prompt to determine if users belong to a given information community. From the output, User A belongs, and User B does not.}
\label{tab:expansion_prompt}
\end{center}

\vspace{-20pt}
\end{table}

\subsection{Iterative Community Expansion}
\label{sec:repeat_comm_expand}
The above process in Sec.~\ref{sec:expand_gpt3} of expanding the validated communities can then be repeated until all users are assigned to or rejected for validated communities, defining convergence.

In addition, we also use a subset of rejected users to form a new set of human validated information communities. To do this, we repeat the above process from Sec.~\ref{sec:graph_comms}-Sec.~\ref{sec:expand_gpt3}: cluster rejected users, summarize them, ask humans to form a new validated community, and then expand the validated community. After each iteration, we have an additional pair of human validated communities.

\subsection{Unsupervised Graph Training}
\label{sec:learning_graph_embeddings}
Above, when we form communities, we create new graph edges connecting users in the same community. We now further learn these edges/user relationships, by fine-tuning the graph model from Sec.~\ref{sec:model}, all without using any additional gold labeled data. For this, we train graph link prediction, which captures this new edge knowledge directly, encouraging connected nodes to have similar embeddings. We do it only on the sub-graph of content that was interacted on: the users and the articles/sources they are directly connected to. Specifically, we train connected nodes to be closer together in the embedding space, while user nodes in different communities should be farther apart.

After this training, the graph model captures the knowledge from the user communities identified by our framework, and can thus be  directly used to better classify news sources for profiling. This is because, in the updated graph model, the new user embeddings directly affect the sources, through either direct or indirect edge connections. 

\subsection{Framework Recap}
In short, we aim to build user information communities, used by the graph model for better news media profiling. Our framework first uses graph models to build candidate communities, which are validated by humans. Identifying the communities is hard for LLMs, but simple for humans and can be done in a few minutes. Then, the communities are expanded. The graph model generates candidates, which the LLM can validate, as it has training examples from the human validation, and just has to identify the same user similarity, a much simpler task. The entire process can be done iteratively and rapidly (under 10 minutes for 5 interaction steps). 
 
 \section{Experiments}
 \label{sec:experiments}
 \subsection{Evaluation Settings}
\label{sec:settings}
We evaluate our framework's ability to improve fake news and news source bias detection. We focus on one of the most challenging settings for a graph framework, the fully inductive setting. Here, in addition to test data not being seen at training time, \textbf{all test nodes are not connected in any way to training set nodes}. For ex., users interacting with test set articles do not interact with any sources/articles/users seen at train time. While this setting is particularly difficult, as social media information learned at training time can't be directly used to improve test performance, it can occur, such as when a new bot farm spreads content.

In addition to the inductive setting, we also focus our evaluation on emerging news events, where all test data is from a specific event collected from a time period after the training time period. Not only is this one of the most common real-world applications for fake news source and bias detection, but it is also very challenging, as test data focuses on sub-events not seen at training. In this work, we evaluate two important news events: \textit{Black Lives Matter (BLM)} and \textit{Abortion/Feminism}.

\begin{table}
\begin{center}
\begin{tabular}{|p{2.4cm}|p{0.7cm}|p{0.7cm}|p{0.7cm}|p{0.7cm}|p{0.7cm}|p{0.7cm}|p{0.7cm}|}
  \hline
  {\textbf{\small Model}} & {\textbf{\small Baly Acc.}}
  & {\textbf{\small Baly F1}} & {\textbf{\small Test Acc}} & {\textbf{\small Test F1}} \\

 \hline
  %SVM & & &  & \\
  \small Baly & \small 71.52 & \small 67.25 & - & - \\
  \small Mehta R-GCN & \small 68.90 & \small 63.72 & - & - \\
  \small Mehta BEST & \small 72.55 & \small 66.89 & - & - \\
  \hline
  \small BL: Mehta R-GCN & \small 65.82 & \small 53.19 & \small 41.89 & \small 28.48 \\
  
 \hline
\end{tabular}
\end{center}
\vspace{-10pt}
\captionsetup{justification=centering}
\caption{\small Fake News Source Detection baseline Results on Baly \cite{baly:2020:ACL2020} and the inductive future Black Lives Matter event (Test). Results show that despite achieving high performance on \cite{baly:2020:ACL2020}, the Baseline from \citeauthor{mehta2022tackling} (BL: Mehta R-GCN) struggles in the inductive, emerging news events setting. This baseline is comparable to the state of the art for fake news source detection from \citeauthor{mehta2022tackling} (Mehta BEST) on \cite{baly:2020:ACL2020}.}
\vspace{-15pt}
\label{table:baseline_results}
\end{table}

\begin{table*}[t]
\begin{center}
\begin{tabular}{|p{6.4cm}|p{0.7cm}|p{0.7cm}|p{0.7cm}|p{0.7cm}|p{1.25cm}|p{1.1cm}|p{1.0cm}|}
  \hline
  {\textbf{\small Model}} & {\textbf{\small FN Acc}} & {\textbf{\small FN F1}} & {\textbf{\small Bias Acc}} & {\textbf{\small Bias F1}} & {\textbf{\small \# Users; \newline \# Sources}} &  {\textbf{\small \# Edges}}  & {\textbf{\small \# Interactions}} \\

 \hline
  \small Baseline: \cite{mehta2022tackling} & \small 41.89  & \small 28.48 & \small 46.79 & \small 27.43 & \small - & \small - & \small - \\
  \hline
  \small Graph Only: High Purity 2 Communities (Comms.) & \small 43.01 & \small 28.85  & \small 46.15 & \small 28.59 & \small 25; 25 & \small 1,200 & \small - \\
  \small Graph Only: High Purity 4 Communities (Comms.) & \small 41.89 & \small 27.23 & \small 48.71 & \small 21.83 & \small - & \small - & \small - \\
  \hline 
  \small LLM Only: 2 Comms, 2 Expansion Rounds & \small 42.70 & \small 28.05 & \small 45.01 & \small 27.84 & \small 38; 63 & \small 494 & \small - \\
  \small LLM Only: 4 Comms, 2 Expansion Rounds & \small 42.70 & \small 28.62 & \small 39.50 & \small 33.22 & \small 69; 56 & \small 1,791 & \small - \\
  \small LLM Only: 6 Comms, 2 Expansion Rounds & \small 40.54 & \small 26.88 & \small 37.03 & \small 29.22 & \small 73; 63 & \small 1,612 & \small - \\
  \hline 
  
  \small LLM + Humans: 2 Comms, 2 Expansion Rounds & \small \textbf{52.51} & \small \textbf{38.03} & \small 44.23 & \small 33.40 & \small 25; 26  & \small 367 & \small 1 \\
  \small LLM + Humans: 2 Comms, 4 Expansion Rounds & \small 46.36 & \small 35.03  & \small \textbf{49.35} & \small \textbf{45.13} & \small 72; 56 & \small 1,087 & \small 1 \\
  \small LLM + Humans: 4 Comms, 2 Expansion Rounds & \small 43.01 & \small 32.36 & \small 47.43 & \small 32.00 & \small 55; 43 & \small 808 & \small 2 \\
  \small LLM + Humans: 6 Comms, 2 Expansion Rounds & \small 41.34 & \small 32.36 & \small 48.07 & \small 33.91 & \small 82; 61  & \small 1,696 & \small 3 \\
 \hline
\end{tabular}
\end{center}
\vspace{-10pt}
\captionsetup{justification=centering}
\caption{\small Fake News (FN) and Bias Source Detection on Black Lives Matter: We evaluate Test Set Accuracy, Macro F1, the \# of users and sources directly connected, the \# of edges created, and the \# of human interactions are performed (each forms 2 validated communities). Results show that our proposed approach, the human interaction models (LLM + Humans, last group), achieve improvements over all other models in Acc. and/or F1. Also, human interactions are critical, as LLM Only models (third group; they still use graph insight) do not achieve significant improvements over baselines (first and second group). Moreover, our best performance is with only 1 single human interaction, creating 2 communities and then expanding them (2 expansion rounds for fake news source detection and 4 for bias source detection).}
\label{table:blm_fake_news_source}
\vspace{-5pt}
\end{table*}

\subsection{Data Collection and Usage}
\label{sec:dataset}
\textbf{Fake News and Bias Source Detection}:  In order to evaluate our framework's ability to improve fake news and news source bias detection, we used the Media Bias/Fact Check dataset, originally collected by \citeauthor{baly:2018:EMNLP2018}. As we focus on specific events, many of which have occurred since the dataset was originally collected, we expand it by scraping additional news sources from Media Bias/Fact Check\footnote{\url{https://mediabiasfactcheck.com}}. Additionally, we scraped the data used to construct the graph in Sec.~\ref{sec:model} (articles sources publish, Twitter users, Twitter interactions, etc.) following the process in \citeauthor{mehta2022tackling}. As done in \citeauthor{baly:2018:EMNLP2018}, we label news sources on a 3-point factuality and 3-point bias scale: \textit{high}, \textit{mixed}, or \textit{low} factuality and \textit{left}, \textit{center}, or \textit{right} bias. Dataset details, including statistics for number of sources is in App.~\ref{sec:dataset_details} and Tab.~\ref{table:data_statistics}. Our code and anonymized data is available.\footnote{\url{https://github.com/hockeybro12/Interactive_News_Media_Profiling}}

\textbf{Events:} For each event that we tested on (Black Lives Matter and Abortion/Feminism), we scraped data for 2 different time periods (01/02/2019 -  06/01/19; 06/02/19 - 05/06/22), searching relevant hashtags on Twitter. These time periods also cover a broad range of sub-events, allowing us to test how our models would do on emerging news events. To learn the graph model for fake news and bias source detection from Sec.~\ref{sec:model}, we used the first period  and a subset of data from \citeauthor{baly:2020:ACL2020} (training the model on the event and general news). The other time period is our test data, and forms a fully inductive graph, where none of the nodes in the test graph are connected to training set nodes, making it hard.

\subsection{Evaluation}
\label{sec:evaluation}
We evaluate our models primarily on Accuracy and Macro F1 score (the dataset is unbalanced), for sources. We also evaluate the \# of users and sources interacted on, the total \# of edges added by all interactions, the \# of expansion rounds done (defined in Sec~\ref{sec:expand_gpt3}), and the \# of interactions done.

\subsection{Baselines}
\label{sec:baselines}
Our first baseline is the strong graph based fake news source detection model from \citeauthor{mehta2022tackling}, which we also trained for and evaluated on bias detection. They also compared to multiple baselines in their work. Tab.~\ref{table:baseline_results} shows the performance of this model on \citeauthor{baly:2020:ACL2020}, but when evaluated on BLM in the inductive setting, it struggles (Tab.~\ref{table:baseline_results}, Tab.~\ref{table:blm_fake_news_source}). 

Our second baseline is our information community detection approach without humans and LLMs, creating  the communities based only on graph model embeddings (Graph Only). We k-means cluster user embeddings, and choose high purity clusters, keeping the top $m$ similar users. We choose k=35 based on validation set performance.

Our final baseline, LLM Only, is our framework without humans, but still using LLM + graph knowledge. To remove the interaction step from Sec.~\ref{sec:inital_human_interaction}. where humans form validated communities by reading summaries and Chat-GPT's assignments, we instead trust the Chat-GPT assignments and use these as the ``validated'' communities.

\subsection{Interactive Framework Results}
\label{sec:interaction_results}
Results for Black Lives Matter fake news and bias source detection are in Tab.~\ref{table:blm_fake_news_source}. Abortion/Feminism results are in Tab.~\ref{table:abortion_fake_news_source} + Tab.~\ref{table:abortion_fake_news_bias}. Results show how our interactive framework (LLM + Humans) enables minimal human interactions (details about interaction process in App.~\ref{sec:human_interaction_details}), sometimes only one, to lead to performance improvements for these tasks, even on emerging news events without additional labels. We experimented with a varying \# of validated communities and expansion rounds (using the dev. set to find the \#), and all showed improvements in either Acc. or F1 score over baselines. Specifically, we see $\sim$33\% improvement on fake news source macro F1, and $\sim$40\% improvement on bias news source macro F1. Our best models for each task and each event only needed up to two human interactions, showing the benefit of our framework to amplify human interactions. Also, all human interaction models outperform all non-human baselines, including LLM Only, showing that both LLM and human insight (to sort out LLM inconsistencies) is critical for news media profiling. 

In summary, these results shows how we are able to successfully decompose the task of finding information communities: taking advantage of graph, LLM, and human strengths, to successfully profile news media, even on emerging news events. 
 
 \section{Discussion}
 \label{sec:discussion}
 In this section, we evaluate our Black Lives Matter interactive framework (Sec.~\ref{sec:interaction_results}) learned information communities. We begin by analyzing the cohesiveness of the communities, first human  (Sec..~\ref{sec:human_characterize}) and second automatically (App.~\ref{sec:community_cohesiveness}). We then show why human interactions are critical (Sec.~\ref{sec:importance_of_humans}). Finally, we analyze the communities themselves, analyzing the topics discussed (App.~\ref{sec:community_analysis}).

\subsection{Human Interactor Analysis}
\label{sec:human_characterize}
In this section, we manually analyze our human interaction process, by asking the interactor how many candidate users for each information community they used and did not use to represent it, in each human community validation round. Results in Tab.~\ref{tab:human_characterize_comms} show that as more interaction steps occur, the candidate users become more similar, as humans reject less users. This shows how our interactive process improves the model's understanding of the social media framework.

\begin{table}[ht!]
\begin{center}
\begin{tabular}{|c|c|c|}
  \hline
  {\textbf{\small Interaction Round}} & {\small \textbf{Users Accepted}} & {\small \textbf{Users Rejected}}\\
 \hline
  \small 1 & \small 3 & \small 9  \\ 
  \small 2 & \small 3 & \small 6  \\ 
  \small 3 & \small 4 & \small 2  \\ 
  \small 4 & \small 6 & \small 1  \\ 
 \hline
\end{tabular}
\caption{\small The number of users accepted and rejected by human interactors in each new community creation step. As more interactions occur, the \# of rejected users decreases, as the graph model learns to better capture similarity. Note that the \# of users presented to humans changes based on cluster sizes.}
\label{tab:human_characterize_comms}
\vspace{-5pt}
\end{center}

\vspace{-20pt}
\end{table}

\subsection{Importance of Humans}
\label{sec:importance_of_humans}
As the results in Sec.~\ref{sec:interaction_results} show, the human interaction step is critical to improve news media profiling performance. This is because LLM's (i.e. Chat-GPT) cannot accurately capture user similarity, particularly for new news events, which leads to non-cohesive communities. However, humans with general world knowledge can easily determine this. As an ex. in Fig.~\ref{fig:chat_gpt_output} (more in App.~\ref{sec:llm_failures}), Chat-GPT responds vaguely that all users share the same perspective, when User 3 is clearly more hostile. 

Without cohesive initial communities, the training examples used to prompt the generation of further communities in Sec.~\ref{sec:expand_gpt3} will also be non-cohesive and thus incorrect, leading to non-cohesive expanded communities. Thus, the graph model wouldn't gain any insight about user perspectives through the communities, which is why downstream performance doesn't improve.

 \section{Conclusion}
 \label{sec:summary}
 In this paper, we proposed a framework for interactive news media source profiling. Our framework combines the strengths of graph based news media profiling models, LLMs, and humans, to build stronger information communities. We show how without any additional labeled data, and less than 5 human interactions which can be done in under 10 minutes, we can better detect fake news and bias sources, on two separate news events, even in the most challenging setting of emerging news. 

Our future work is building larger and better communities, and having more human interaction rounds. We hypothesize that more data (users, sources, articles, and human interactions) could lead to lead to better communities, as our approach can capture more perspectives on social media. This would also likely lead to a better trend in the results, leading to more consistent performance improvements as the number of interaction rounds are increased.

 \section{Acknowledgements}
 We thank the anonymous reviewers of this paper
for their vital feedback. The project was partially funded by NSF award IIS-2135573 and NSF CAREER award IIS-2048001. 

 \section{Ethics Statement}
 For the ethics statement, we first discuss limitations of our model (\ref{sec:limitations}), and then in Sec.~\ref{sec:ethics} we discuss ethics for deploying our models..
 
 \subsection{Limitations}
 \label{sec:limitations}
 In this paper, we focus on news media profiling (fake news and bias source detection) on English and Twitter, specifically in the Black Lives Matter domain. The experimental results we presented in this paper showed our framework works in these domains/tasks. We are hopeful and believe that our framework would generalize to other domains, tasks, and topics, but we leave the investigation of this to future works.

 In this paper, we also primarily focused on the evaluation setting of early detection of fake/biased news sources, where we evaluate on unseen test data that is not connected to any training set data in the graph. We believe that this is one of the most challenging settings for news media profiling, as shown by prior work. We thus believe that our framework would generalize to other news media profiling settings, including ones that are not in the early detection space. Our future work involves testing this hypothesis, by combining our frameworks with other works in the early detection space.

 Our framework utilizes Large Language Models, specifically GPT-3, which the details are not yet fully known publicly. Although these models have been shown to achieve strong performance in numerous NLP benchmarks  \cite{qinChatGPT}, we believe the community should still be careful in deploying them.

 Our framework also utilizes human interactions, which in our paper are extremely simple, as humans must just read short summaries to determine similarity. Further, our framework needs an extremely small amount of human interactions. However, we still caution that in a real world deployment of our framework, we should be careful of human interactors and make sure they do not have a malicious intent and are well educated for this task. Moreover, it would be better if numerous humans provided judgements on a single interaction sample, to confirm all the interactions across multiple experts.
 
 For our experiments, we used a single GeForce GTX 1080 NVIDIA GPU, with 12 GB of memory. As our models are largely textual based, they do not require much GPU usage, but this could change in real world settings, where lots more data is available, which could be a potential limitation. Our hyper-parameter search, mentioned was done manually, based on dev set performance. The appendix provides more model details.
 
 \subsection{Ethics}
 \label{sec:ethics}
 To the best of our knowledge, we did not violate any code of ethics throughout the experiments done in this paper. We reported technical details necessary to reproduce our results, and will release the code and dataset we collected, upon publication. We evaluated our model on the datasets that we collected in this paper, and was collected by prior work, but it is possible that results may differ on other datasets. However, we believe our methodology is solid and applies to any social media news profiling setting, as shown by our performance on emerging news events. 
 
 Due to lack of space, we placed some of the technical details in the Appendix section. The results we reported support our claims in this paper and we believe that they are reproducible. Any qualitative result we report is an outcome from a machine learning model that does not represent the authors' personal views. 
 
 In our future dataset release, we include sources, users, and articles, so that our experiments can be replicated. Each are in English, and are public information. We map each to an ID, for anonymity, and release Article textual representations. Article texts are available for academic use, and can be provided by requesting the authors and agreeing to appropriate conditions.
 
 Our framework in general is intended to be used to profile news media sources, and help identify the spread of misleading or perspective changing content on social media. While our framework could be used to build better methods of avoiding fake news/bias detection by ML systems, our interactive framework can guard against that as well. 
 
 In general, we caution that our models and methods be considered and used carefully, as in an area like news media profiling there are great consequences of wrong model decisions, such as unfair censorship and other social related issues. Further, it is possible our models are biased, and this should also be taken into consideration. An important future work is to investigate our models, interpreting them and understanding their predictions even better than the analysis showed in the Discussion section of this paper. 

The interactive setting we proposed was successful in this paper, particularly because the interactions were simple. However, in the real world, there could be biased interactors with malicious motives, and that is an important thing to consider when dealing with fake/bias news source detection systems.

 These and many other related issues are things to consider when using models such as the ones proposed in this work.

\bibliography{anthology,custom}

\newpage
\appendix

\section{Experimental Settings}
\label{sec:experimental_settings}

\subsection{Profiling News Media Sources Task Definition and Importance}
\label{sec:source_task_importance}
In this paper, we focused on detecting the political bias and factuality of news media sources, which we call \textbf{news media profiling}. We focused on these tasks as stopping misinformation is critical, and politically biased content can sway beliefs and affect important real-world events, such as political elections \cite{vosoughi2018spread}.

We model sources for factuality on a 3-point scale: \textit{low}, \textit{mixed}, and \textit{high}. Similarly, we model sources for political bias on a 3-point scale: \textit{left}, \textit{center}, and \textit{right}. More details about this task setup and its importance can be found in \cite{baly:2020:ACL2020}. Below, we also briefly discuss the importance of profiling the source itself.

Focusing on profiling the source itself, rather than the content, can have several benefits, which is why we focused on this task work: \textbf{(1)} Most sources publish a large amount of content, and knowing the facutality/bias of the source can give us insight about all the content they publish. Moreover, any new content that the source publishes is more likely to have a similar factuality/bias as the source's historical content. Thus, knowing the source's historical level of facutality/bias can provide insight about the new content, which can help in rapidly profiling it. For example, a source publishing mostly \textit{left} biased content in the past is more likely to be \textit{left} biased in the future. \textbf{(2)} There are a lot more sources than content on social media, so it can be easier to accurately profile sources, and our framework makes this task even easier. As mentioned above, doing this can then provide insights on all the content the source publishes. \textbf{(3)} There are many new news sources arising daily, so manually profiling all of them is extremely difficult. Thus, developing an automated system is critical.

\subsection{Graph Initial Embeddings}
\label{subsec:initial_embeddings}
We followed the released code and data from \citeauthor{mehta2022tackling}, so we use their exact node embedding representations as our initial graph embeddings. The Twitter embedding is a 773 dimensional vector consisting of the SBERT \cite{reimers2019sentence} RoBERTa \cite{liu2019roberta} representation of the user profile, consisting of features such as: user bio, user verification status, number of user followers/following, how many tweets they post, and how many likes their tweets have received. We also used similar features for YouTube embeddings for each source: number of likes, dislikes, and number of comments on their videos. For the article feature vector, it was also a SBERT RoBERTa textual embedding.

\subsection{Models and Training}
\label{appendix:model_training}
We used the ChatGPT models available via the OpenAI API\footnote{\url{https://openai.com/blog/openai-api}} as of February 2023. 

For our graph models, we used the publicly released code and hyper-parameters of \citeauthor{mehta2022tackling}, which uses the PyTorch \cite{NEURIPS2019_9015} and DGL (Deep Graph Library) \cite{wang2019deep} libraries in Python. The R-GCN has 5 layers, 128 hidden unites, a batch size of 128, and learning rate 0.001. 

Our models are trained using a 12GB Titan XP GPU card, and intial training takes 2 hours. Link prediction training after human interactions is very quick, and can complete in under 30 minutes. Further, doing the expansion step completes in under 30 minutes. It took under 10 minutes for the humans to do all the interactions.

We used the development set to evaluate model performance, and choose the best hyper-parameters for our experiments.

Our models are trained for source classification, using a separate Fully Connected (FC) layer for each fake news source classification and political bias source classification. The R-GCN \cite{schlichtkrull2018modeling} model we use creates contextualized graph embeddings for each node in the graph. For example, source embeddings are affected by users and articles they are directly or indirectly connected to. This is why our approach to learn user communities, which leads to better user embeddings (i.e. users with similar perspectives are closer together), leads to better source embeddings.

After being passed through the FC layer for classification, the R-GCN source embeddings are then passed through the Softmax activation function, and finally used to predict the source label. The model is trained like in \citeauthor{mehta2022tackling}, using a categorical cross-entropy loss, where the gold training labels are factuality or political bias. The two source classification tasks are trained jointly, summing their losses.

\section{Prompts for GPT-3}
\begin{figure*}[t!]
  \centering
  \includegraphics[scale=0.5]{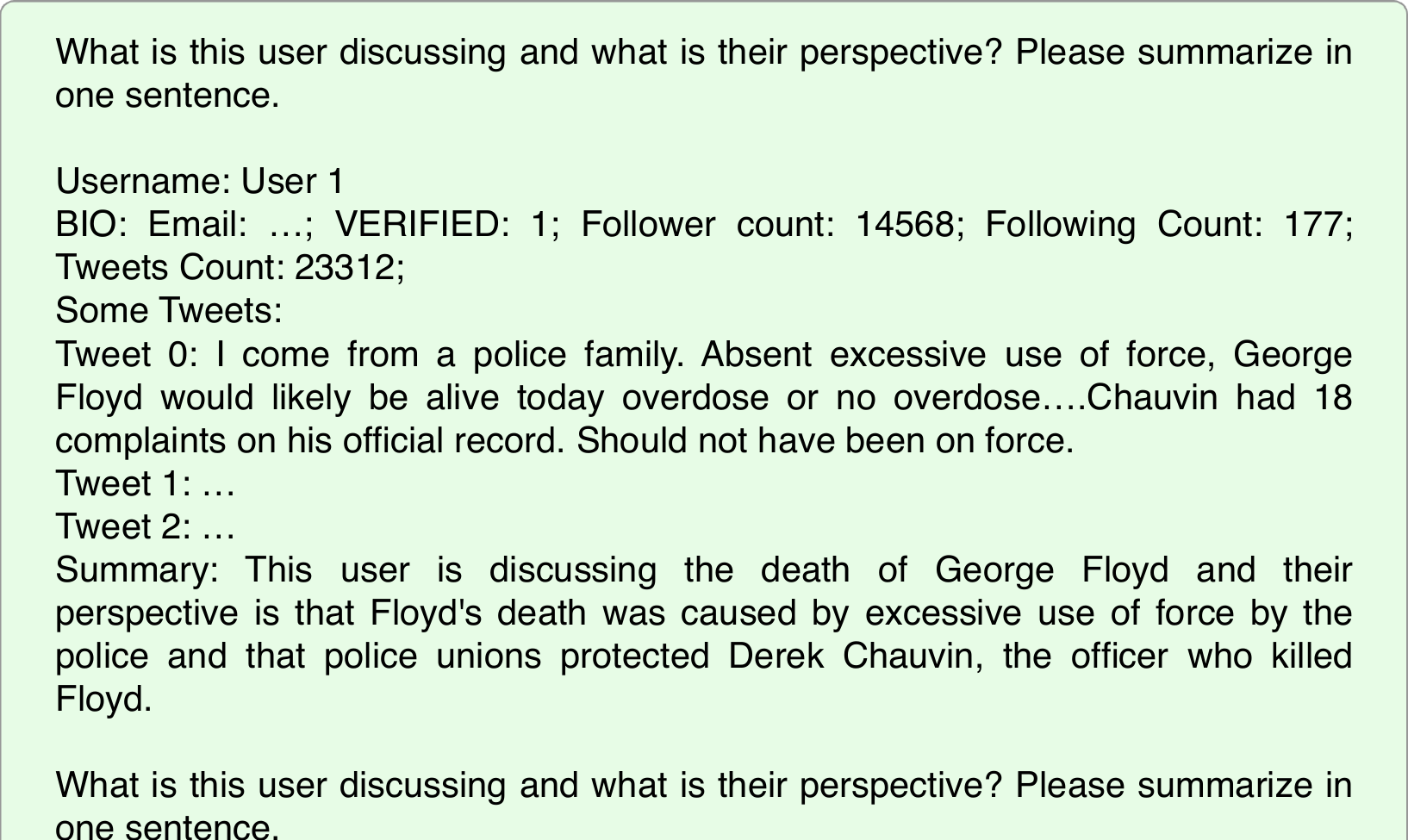}
\caption{\small An example of the prompt we used to determine the user summary. Based on their bio, meta-data, and tweets, we create a summary.}
\label{fig:user_summary}
\vspace{-10pt}
\end{figure*}

\begin{figure*}[t!]
  \centering
  \includegraphics[scale=0.5]{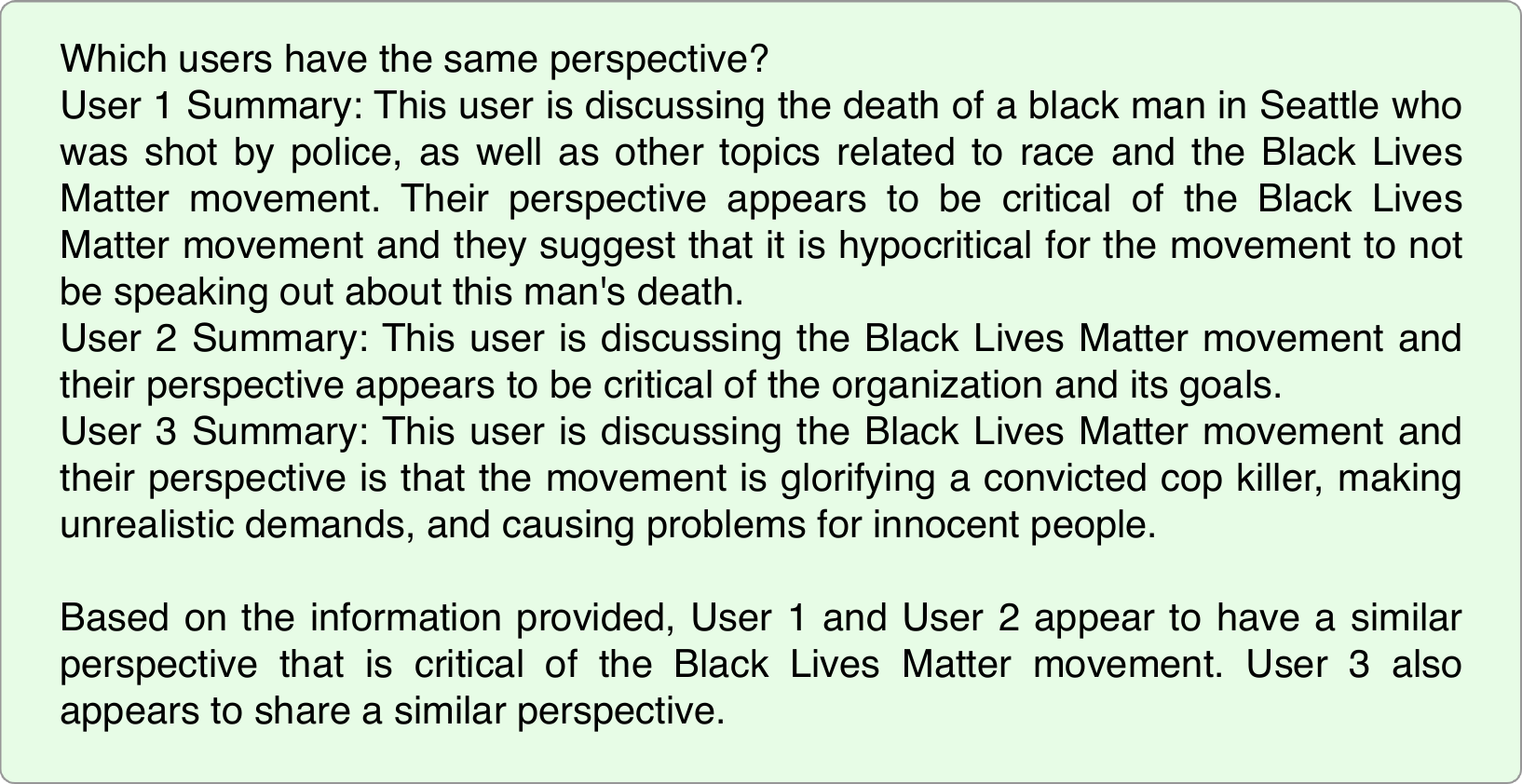}
\caption{\small An example of the output shown by Chat-GPT when provided user summaries and asked to predict similarity. Note how often times the output can be vague, which is why human interactions are necessary.}
\label{fig:chat_gpt_output}
\vspace{-10pt}
\end{figure*}

\begin{figure*}[t!]
  \centering
  \includegraphics[scale=0.5]{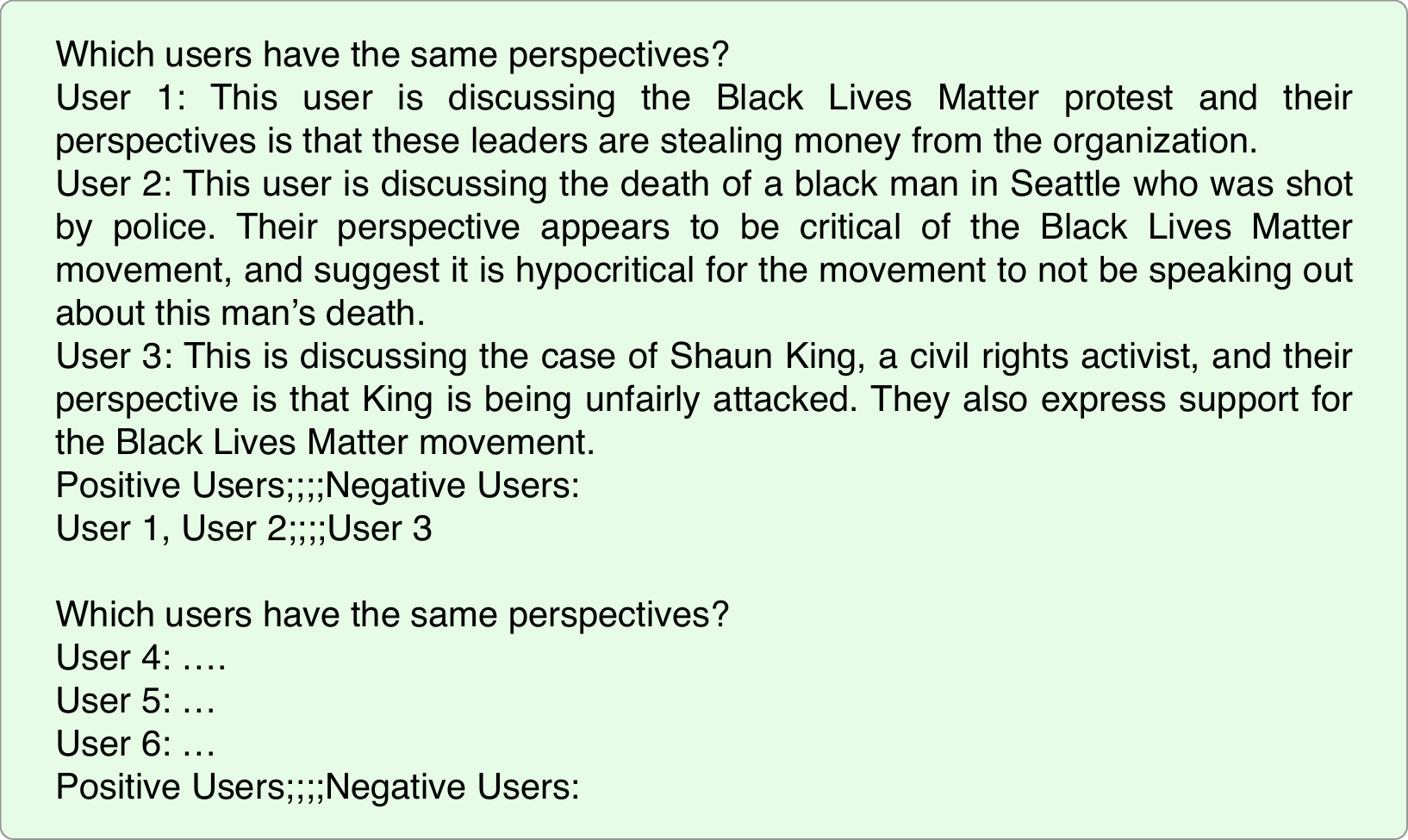}
\caption{\small An example of the prompt we used to determine community membership for one of the human validated information communities. We use the first paragraph as a 1-shot example, to prompt the model. User 1 and 2 are both critical of the Black Lives Matter movement protests, and thus part of the same community, while User 3 is in support of it, and thus shouldn't be in the community. Based on this, we prompt GPT-3 with additional users (in this case User 4, 5, and 6), and ask it to determine which users belong in the community and which do not.}
\label{fig:user_comm_membership}
\vspace{-10pt}
\end{figure*}

In this section, we describe the prompts we use for GPT-3 and Chat-GPT, to utilize the large amounts of pre-trained knowledge contained in these models to help us in our interactive framework. It has been shown \cite{brown2020language, weichain} that prompting can help utilize LLMs like GPT-3 for many NLP tasks, as they cannot be trained directly. While these models cannot do complex reasoning to determine if sources are fake or biased news, they can solve simpler tasks, such as determining if users have the same perspectives, which can help with our tasks, that we built into our interactive framework. Specifically, they can summarize users based on their content (Fig.\ref{fig:user_summary}), help humans analyze perspectives (Fig.~\ref{fig:chat_gpt_output}), and determine community membership based on similarity (Fig.~\ref{fig:user_comm_membership}),

All the prompts used in this paper are human designed.

\section{Human Interaction Details}
\label{sec:human_interaction_details}
We used a single human interactor for all our experiments, who was a computer science PhD student, of Asian American descent. The student was compensated in research credit hours, as the interaction process was part of their research credit work.

We used only a single human due to the fact that our interaction process is very simple, as humans only read a few user summaries and determine content similarity, which in all cases we evaluated on is fairly straight-forward. Moreover, the amount of interactions done in this work was very small, as only 5 interactions could lead to the formation of 8 information communities and connect different communities together. The human we used was also expertly trained. Thus, we did not explore using additional interactors in this work, and it is something we leave for future work. For example, other setups could use multiple experts for human interactions, taking their majority vote as the final interaction. 

\section{Dataset Statistics}
\label{sec:dataset_details}
Table~\ref{table:data_statistics} shows the statistics of the number of sources for the Black Lives Matter and Abortion/Feminism event we evaluate on. 

To collect the data for Black Lives Matter and Abortion/Feminism, we searched hashtages on Twitter. The hashtags/search terms we used for the \textit{Black Lives Matter} event were: \textit{Black Lives Matter, BLM, blacklivesmatter, Floyd, George Floyd}. The hashtags/search terms we used for the \textit{Abortion/Feminism} event were: \textit{abortion, feminism, womenempowerment, womensrights, metoo, prolife, prochoice}.

\begin{table*}[t]
\begin{center}
\begin{tabular}{|p{4.0cm}|p{2.5cm}|p{2.5cm}|p{2.5cm}|}
  \hline
  {\textbf{\small Dataset}} & {\textbf{\small Low Factuality or Left Bias}} & {\textbf{\small Mixed Factuality or Center Bias}} & {\textbf{\small High Factuality or Right Bias}} \\

 \hline
  \small Black Lives Matter Bias & \small 49 &  \small 56  &   \small 74 \\
  \small Black Lives Matter Factuality & \small 35 &  \small 45  &   \small 76 \\
  \hline
  \small Abortion + Feminism Bias & \small 38 &  \small 50 &  \small 91 \\
  \small Abortion + Feminism Factuality & \small 49 & \small 72 & \small 82 \\
  \hline 
\end{tabular}
\end{center}
\vspace{-10pt}
\captionsetup{justification=centering}
\caption{Number of sources in our datasets for each emerging news event we evaluate on.}
\vspace{-10pt}
\label{table:data_statistics}
\end{table*}

\begin{table*}[t]
\begin{center}
\begin{tabular}{|p{7.5cm}|p{0.8cm}|p{0.8cm}|p{1.3cm}|p{1.2cm}|p{1.1cm}|}
  \hline
  {\textbf{\small Model}} & {\textbf{\small Test Acc}} & {\textbf{\small Test F1}} & {\textbf{\small \# Users; \# Sources}} &  {\textbf{\small \# Edges}}  & {\textbf{\small \# Interactions}} \\

 \hline
  \small Baseline: \cite{mehta2022tackling} & \small 36.04  & \small 23.32 & \small -  & \small - & \small - \\
  \hline 
  \small Graph Only: High Purity 4 Communities & \small 34.95 & \small  19.09 & \small 22; 11 & \small 176 & \small - \\
  \hline 
  \small LLM + Humans: 2 Validated Comms, 2 Expansion Rounds & \small \textbf{37.37} & \small \textbf{25.24} & \small 173; 16 & \small 280 & \small 1 \\
  \small LLM + Humans: 4 Validated Comms, 2 Expansion Rounds & \small 35.43  & \small 19.64 & \small 50; 32 & \small 628 & \small 2 \\
 \hline
\end{tabular}
\end{center}
\captionsetup{justification=centering}
\caption{\small Fake News Source Detection on Abortion/Feminism: We evaluate Test Set Accuracy, Macro F1, the number of users and sources directly connected/connected to, the number of edges created, and how many human interactions are performed. Results show both of our human interaction models (LLM + Humans) achieve improvements over other models (Baseline and Graph Only model). Specifically, creating two human validated communities and then expanding them over 2 expansion rounds achieves the highest fake news source detection performance. The final communities have 173 users, interact directly with 16 sources, and create 280 edges. Moreover, this performance improvement is with only 1 single human interaction.}
\label{table:abortion_fake_news_source}
\vspace{-10pt}
\end{table*}

\begin{table*}[t]
\begin{center}
\begin{tabular}{|p{7.5cm}|p{0.8cm}|p{0.8cm}|p{1.3cm}|p{1.2cm}|p{1.1cm}|}
  \hline
  {\textbf{\small Model}} & {\textbf{\small Test Acc}} & {\textbf{\small Test F1}} & {\textbf{\small \# Users; \# Sources}} &  {\textbf{\small \# Edges}} & {\textbf{\small \# Interactions}}  \\

 \hline
  \small Baseline: \cite{mehta2022tackling} & \small 46.92  & \small 33.21 & \small -  & \small - & \small - \\
  \hline 
  \small Graph Only: High Purity 4 Communities & \small 47.86 & \small  34.02 & \small 22; 11 & \small 176 & \small - \\
  \hline 
  \small LLM + Humans: 2 Validated Comms, 2 Expansion Rounds. & \small 46.92 & \small \textbf{40.33} & \small 173; 16  & \small 280 & \small 1 \\
  \small LLM + Humans: 4 Validated Comms, 2 Expansion Rounds & \small \textbf{51.39}  & \small 38.05 & \small 50; 32 & \small 628 & \small 2 \\
 \hline
\end{tabular}
\end{center}
\captionsetup{justification=centering}
\caption{\small Bias News Source Detection on Abortion/Feminism: We evaluate Test Set Accuracy, Macro F1, the number of users and sources directly connected/connected to, the number of edges created, and how many human interactions are performed. Results show both of our human interaction models (LLM + Humans) achieve improvements over other models (Baselines and Graph Only). Specifically, creating four human validated communities and then expanding them over 2 expansion rounds achieves the highest bias news source detection accuracy. The final communities have 50 users, interact directly with 32 sources, and create 628 edges. Moreover, this performance improvement is with only 2  human interactions.}
\label{table:abortion_fake_news_bias}
\end{table*}

\section{Related Work Cont.}
\label{appendix:related_cont}
We now discuss additional related works we didn't cover in the main paper, due to space. Several additional works aim to analyze fake news spreaders on social media, some of them using graphs \cite{rath2020detecting, rath2021scarlet, sakketou2022temporal}.

\subsection{Impact of Information Communities for News Media Profiling}
Prior work has shown that misinformation tends to spread in groups on social media  \cite{bessi2016homophily, halberstam2016homophily, cinelli2021echo}. Specifically, they show that like-minded users tend to form groups, that biased/false information reaches the users in these groups more quickly, and that these groups are more likely to be biased/spread misinformation. 

This motivates our ideas to build better information communities for fake news and bias source detection. We hypothesize that if we can identify these like-minded users, i.e. our information communities, then we can more easily identify their users' bias and likeliness to spread misinformation. This knowledge can then be used to profile news media sources better, especially through our graph model, where users, articles, and sources are directly connected and vary in similarity to each other. Thus, we aimed to build better information communities, and took advantage of minimal human interactions to do this.

\subsection{Humans Interacting with LLMs}
Humans interacting with LLMs has gained popularity recently. One direction is Reinforcement Learning for Human Feedback (RLHF) \cite{bai2022training, OpenAI2023GPT4TR}, where humans preferences are used to train a Reinforcement Learning Agent reward model, which can then be used to improve the LLM. An extension of this, which needs significantly less human interactions, is having humans provide a few training examples to a LLM prompt, which can be used by the LLM to generate rewards \cite{kwonreward}. In another direction, human interactions can be used to generate LLM prompt instructions to better solve a variety of NLP Tasks \cite{zhang2023wisdom}. 

In contrast, in this work, we had humans interact with LLM and graph knowledge to build better information communities for news media profiling. To do this, we first used LLMs to generate user and community summaries, which we presented to humans. Humans were then asked to use their judgement and reasoning skills, something that is simple for them but hard for LLMs, to form initial validated communities. These validated communities consisted of users who had similar perspectives on similar topics. Then, using graph knowledge, we generated additional candidate communities, and asked LLMs if the users in these additional communities belonged to any of the the human validated communities. As a prompt for this decision, we used the human validated community assignments. 

In summary, we used LLMs to help humans (i.e. generate summaries) and amplify human interactions (i.e. determine if additional users are similar to users in any human validated communities). 

\section{Discussion Continued}
We now continue our discussion from Sec.~\ref{sec:discussion}, analyzing our Black Lives Matter human interaction models.

\subsection{Community Cohesiveness Analysis}
\label{sec:community_cohesiveness}
\begin{table*}[ht!]
\begin{center}
\begin{tabular}{|p{0.9cm}|p{2.5cm}|p{4.5cm}|p{4.5cm}|}
  \hline
  {\textbf{\small Comm. \#}} & {\small \textbf{Dominant LLM + Humans Label}} & {\small \textbf{LLM Only: \% Of Users with Dominant Label}} & {\small \textbf{LLM + Humans: \% Of Users with Dominant Label}}\\
 \hline
  \small 1 & \small Right &  \small $\sim$50\% & \small $\sim$60\%  \\ 
  \small 2 & \small Right  & \small $\sim$37\% & \small $\sim$58\%\\ 
  \small 3 & \small Right & \small $\sim$43\% & \small 100\% \\ 
  \small 4 & \small Center & \small 40\% & \small 50\% \\ 
  \small 5 & \small Left & \small $\sim$71\% & \small $\sim$66\% \\ 
 \hline
\end{tabular}
\caption{\small At the final expansion round (i.e. after multiple steps of human interaction + model expansion) the majority of each community's users for the LLM + Human Interaction Model (last column) have the same gold bias label as the dominant one in the community, showing high cohesiveness (at least in gold bias label). On the contrary, the LLM Only model (third column) has a lower percentage of users with the same gold bias label, showing that without human interactions it is harder to learn user perspectives, at least based on this approximation analysis.}
\label{tab:community_cohesiveness}
\end{center}
\end{table*}

In this sub-section, we automatically analyze how many users in each community have the same perspectives. To do this, as an approximation, we hypothesize that the communities of users with similar perspectives likely have users with the same bias label. We use bias as an \textit{approximation} as we have gold data for it, and users with the same political bias likely have similar perspectives (i.e. right bias users likely want to lower taxes). Tab.~\ref{tab:community_cohesiveness} shows that even in the final expansion round (i.e. after multiple steps of human interaction + model expansion - LLM + Humans Model) users in the communities largely have the same bias label, both when chosen by humans and automatically expanded. Thus, this approximation shows that our communities are in some ways cohesive. 

On the contrary, in Tab.~\ref{tab:community_cohesiveness}, the LLM Only model doesn't have as many users having the same labels as the LLM + Humans model, showing that without human interactions it may be harder to learn user perspectives.

\subsection{Human Analysis of Community Topics}
\label{sec:community_analysis}
We now manually analyze the information communities learned by our best performing model on the Black Lives Matter event, by looking at the top 5 user summaries, determined by user embedding similarity to the community centroid. We observe that our communities capture meaningful perspectives. One community is against the Black Lives Matter protests, believing they are causing damage and the leaders are not condemning it. Another is in support of them, as they feel police do not treat everyone fairly. Other important sub-topics are also discussed, such as George Floyd murder, Ahmaud Arbery murder, police brutality and unions. All these are important BLM related topics.

\subsection{LLM Only Failure Cases}
\label{sec:llm_failures}
\begin{figure*}[t!]
  \centering
  \includegraphics[scale=0.5]{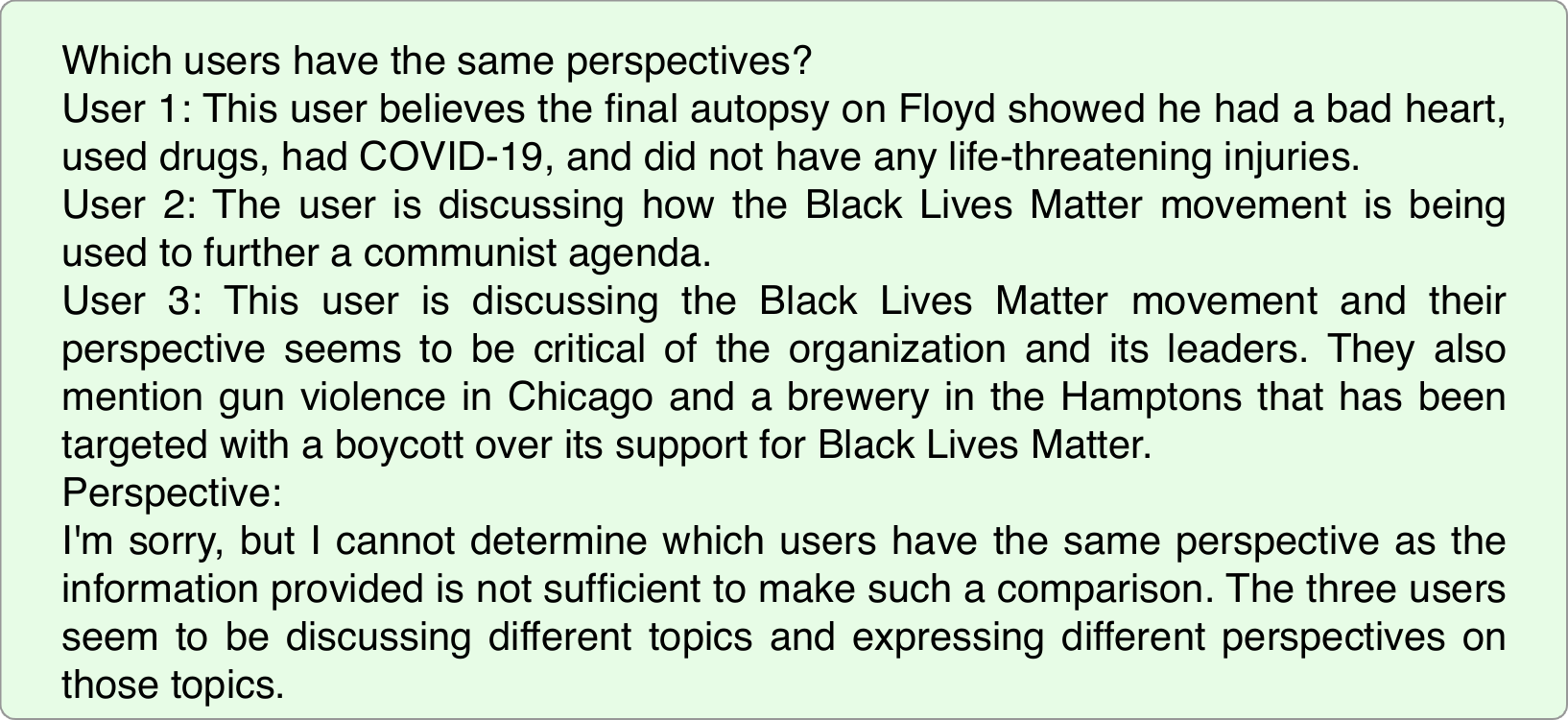}
\caption{LLM Failure Case: \small In this case, the LLM (Chat-GPT) can't find any communities, but it is clear that at least User 2 and User 3 should be in the same community, as they are both against the Black Lives Matter movement.}
\label{fig:llm_failure_blm}
\vspace{-10pt}
\end{figure*}

\begin{figure*}[t!]
  \centering
  \includegraphics[scale=0.5]{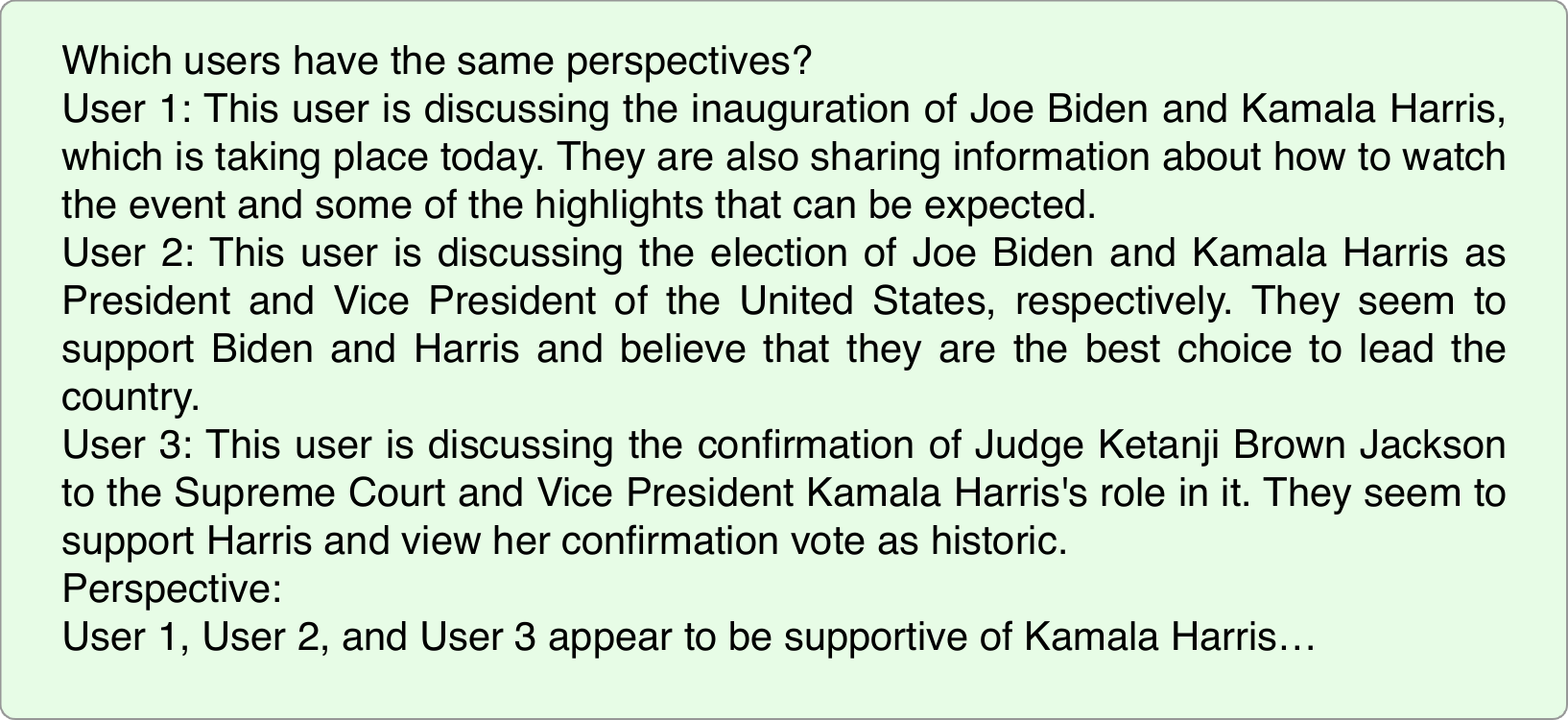}
\caption{LLM Failure Case: \small In this case, the LLM (Chat-GPT) considers all 3 users as similar and in the same community. However, User 1 doesn't belong, as they are likely just reporting on the news of the inauguration. The LLM gets confused over the word ``highlights'', which could be positive or negative in this situation.}
\label{fig:llm_failure_kamala}
\vspace{-10pt}
\end{figure*}

In this section, we provide a few more examples of cases where Chat-GPT couldn't find good information communities, and thus humans were needed, as in Sec.~\ref{sec:human_interaction_details}. The examples are shown in Fig.~\ref{fig:llm_failure_blm} and Fig.~\ref{fig:llm_failure_kamala}, and the captions of the figures describe the failures.

While in this paper we experimented with GPT-3 and Chat-GPT as our LLMs of choice, we hypothesize that our results and framework would hold true for other strong LLMs as well. First, other LLMs are likely to also struggle at finding information communities on unseen data, as it is very challenging for AI models to make inferences on data and topics they have never seen before. Second, it is likely that other LLMs that perform well on determining text similarity can be used with our framework \cite{zhao2023survey}. We primarily used LLMs to expand communities, by asking them to determine if new users are similar to users chosen by humans to be part of validated communities. In some ways, this is a text similarity problem, as user summaries are compared. We leave the further investigation of the choice of LLMs for use with our framework to future work.  

\subsection{Finer Cases}
In this sub-section, we discuss some finer cases that could occur in our community analysis, and how they would or would not affect our framework.

\textbf{Communities spreading both fake news and biased content:} A group of users that spreads both fake news and biased content likely wouldn't affect our approach/results. Through the approach in Sec.~\ref{sec:interactions}, the group would be identified as a single information community. Then, the R-GCN graph model would learn that this community is both a fake news and political bias spreading information community.

% TODO: should we include this?
\textbf{Source Label Inconsistencies:} In this paper, we obtained source labels for both factuality and political bias from Media Bias/Fact Check\footnote{\url{https://mediabiasfactcheck.com}}, a popular source fact-checking and source political bias detection website. This website typically holistically evaluates sources, so it is likely a majority of the content we scraped for our experiments follows the factuality/bias label provided by Media Bias/Fact-Check. However, it is possible that some sources have different levels of factuality/bias for different events that Media Bias/Fact Check doesn't capture (i.e. a source could be labeled as ``mixed'' factuality but be completely factual on Black Lives Matter news). While we leave the analysis of this to future work, we hypothesize that these cases are rare and thus unlikely to significantly affect our results. More importantly, it's possible that our framework can actually make the correct prediction on these incorrectly labeled sources. This can happen as our formed information communities typically span multiple sources and are very cohesive, so they are likely to be accurate. Once we form our communities, we train the graph for link prediction, so we aren't actually using the source labels but rather we are capturing user perspectives through training. Thus, if some sources are labeled incorrectly and that leads to an initially biased graph model, our interaction process can actually fix that by helping us learn a better model, improving performance on those incorrectly labeled sources.

\textbf{Human Interaction Inconsistencies:} As mentioned in the main paper, our interaction task is extremely simple, as humans just have to determine user similarity by reading a few short spans of user text. This distinction is very clear-cut, as users that are borderline similar should not be placed in the same community. Thus, different human interactors are likely to make the same decisions, no matter their backgrounds/beliefs/etc., because identifying this level of similarity is a fairly simple task. Moreover, there are multiple ways to ensure that the interactions are done accurately, such as hiring multiple experts and taking their majority vote. However, still, in this section we analyze the situation in which humans incorrectly choose users as similar when they are not, or vice versa. 

The strength of our framework is forming cohesive information communities through the interactions, where users have similar perspectives. Assuming an interaction lead to a community that wasn't cohesive, it is likely that the trained graph model would learn to ignore it, as it wouldn't gain any significant insight from this ``random'' community. Thus, this ``incorrect'' interaction is not likely to significantly hurt our model. Further, on a large scale over many interactions and lots of communities formed, a few ``incorrect'' interactions is unlikely to make huge negative difference to our approach, due to the fact that we always train for them using link prediction, so the model can learn to ignore it if necessary. For example, even though the link prediction training objective would pull the non-similar users in this ``incorrect'' community closer together, the content these users are connected to in a different ``correct'' community would still be pulled closer together. If there are more ``correct'' communities than incorrect, then even the users in the ``incorrect'' community would be indirectly affected by other communities and end up with the correct representations (i.e. farther apart if they are non-similar).

\end{document}